\documentclass[sigconf]{aamas} 
\usepackage{balance}
\usepackage{algorithm}
\usepackage{algorithmic}

\setcopyright{none}
\renewcommand\footnotetextcopyrightpermission[1]{}
\usepackage{newfloat}
\usepackage{listings}

\title{Leveraging weights signals - Predicting and improving generalizability in reinforcement learning}
\author{Olivier Moulin}
\affiliation{
  \institution{Vrije Universiteit Amsterdam}
  \city{Amsterdam}
  \country{Netherlands}}
\email{o.moulin@vu.nl}

\author{Vincent Francois-Lavet}
\affiliation{
  \institution{Vrije Universiteit Amsterdam}
  \city{Amsterdam}
  \country{Netherlands}}
\email{vincent.francoislavet@vu.nl}

\author{Paul Elbers}
\affiliation{
  \institution{Vrije Universiteit Medical Center Amsterdam}
  \city{Amsterdam}
  \country{Netherlands}}
\email{p.elbers@amsterdamumc.nl}

\author{Mark Hoogendoorn}
\affiliation{
  \institution{Vrije Universiteit Amsterdam}
  \city{Amsterdam}
  \country{Netherlands}}
\email{m.hoogendoorn@vu.nl}

\begin{abstract}
Generalizability of Reinforcement Learning (RL) agents (ability to perform on environments different from the ones they have been trained on) is a key problem as agents have the tendency to overfit to their training environments.
In order to address this problem and offer a solution to increase the generalizability of RL agents, we introduce a new methodology to predict the generalizability score of RL agents based on the internal weights of the agent's neural networks. Using this prediction capability, we propose some changes in the Proximal Policy Optimization (PPO) loss function to boost the generalization score of the agents trained with this upgraded version. Experimental results demonstrate that our improved PPO algorithm yields agents with stronger generalizability compared to the original version.
\end{abstract}


\begin{document}
\pagestyle{fancy}
\fancyhead{}
\maketitle

\section{Introduction}
\label{Introduction}
The ability of reinforcement learning (RL) agents to generalize to previously unseen environments is a critical challenge. Even if RL agents can achieve good performance in environments in which they have been trained, they often struggle to transfer that performance to unseen but similar environments. This overfitting issue has been extensively studied, with contributions from Cobbe et al. \cite{KCobbe2019} and Packer et al. \cite{CPacker2018}, among others. In response, different strategies such as domain randomization or architectural changes have been proposed to address this generalizability issue, for instance, in the works of Moulin et al. \cite{OMoulin2022}, Sonar et al. \cite{asonar2021}, Chen \cite{JZChen2020}, Lu et al. \cite{XLu2020} and Igl et al. \cite{Migl2019}.

This paper takes a different approach: rather than looking for external actions like the ones we just listed, we investigate the structure of an agent's neural network and train a predictor.
This predictor, by identifying patterns in the agent weights can predict the agent generalizability. This predictive capability enables us to introduce a modification to the training process by proposing an enhanced loss function for the agent training algorithm. This change in the loss function explicitly encourages better generalizability.

Our contribution includes:
\begin{itemize}
    \item Build a dataset of Deep RL agents' neural networks, trained on Minigrid and Coinrun environments
    \item Create a predictor to estimate the generalizability of an Deep RL agent based on its Neural Networks' weights 
    \item Incorporate this predictor in the training algorithm loss function to guide the training loop to look not only for performance but also a better generalizability
    \item Achieve better generalizability on agents trained with the improved training algorithm compared to the standard one
\end{itemize}
The remainder of this paper is structured as follows. Section~\ref{related_work} reviews the related works that have a close connection to this area of research. Section \ref{approach} presents the proposed methodology, including the design of the generalizability predictor, the improvements made to the training loop and loss function. Section \ref{experimental_setup} describes the experimental setup used to validate our contributions, while Section \ref{results} presents and analyzes the results of our experiments. Finally, Section \ref{discussion_conclusion}  offers a discussion of key takeaways and directions for future research. \textit{The source code is available on github.com, the link will be included here if selected for publication to preserve anonymity of this review. The full code is also included in the appendix.}

\section{Related work}
\label{related_work}

Our work draws inspiration from the approach introduced by Unterthiner et al. \cite{tUnterthiner2021}, but it diverges both in application and objective. 
In their study, Unterthiner et al. \citep{tUnterthiner2021} demonstrated that it is possible to accurately estimate the test performance of convolutional neural networks (CNNs) on image classification tasks by analyzing their internal weight structures. They were able to reach high predictive accuracy when estimating a model’s performance on a given dataset. They trained a set of CNNs with varying performance levels and used statistical transformations of the neural network weights as input features for predictive models including gradient-boosted machines, and deep neural networks (DNNs). 
Our approach is defined in the context of reinforcement learning (RL) with a focus on predicting generalization score of agents on never seen environments rather than raw performance accuracy. We also take it a step forward by using this generalizability predictor to improve the training algorithm to increase the generalizability of the agents trained with it.

Generalizability in RL has attracted significant attention, with notable contributions from Cobbe et al. \cite{KCobbe2019}, who introduced the CoinRun benchmark to empirically assess generalization gaps, and Zhang et al. \cite{CZhang2018}, who analyzed overfitting in standard RL algorithms. Other work, such as Igl et al. \cite{Migl2019}, Lu et al. \cite{XLu2020}, Chen et al. \cite{JZChen2020} and Schwarz et al. \cite{Jschwarz2018}, have proposed different regularization techniques or structural modifications to improve generalizability. 
Their proposals include incorporating auxiliary losses or information bottlenecks. 
These methods require architectural changes or explicit noise injection during training.
Other proposals like the one from Zhiwei et al. \cite{JZhiwei2022}, Diuk et al. \cite{CDiuk2014} and Justesen et al. \cite{NJustesen2018} focus on adapting the way agents are trained to improve their generalizability.
Our approach does not change the information gathered and interactions between the agent and the environment but focuses on providing a way to increase the generalization by modifying the training algorithm itself, based on signals extracted from the agent neural network's weights.
\section{Approach}
\label{approach}
\subsection{Overall description}
\begin{figure}[!htbp]
	\centering
	\includegraphics[width=4.5cm]{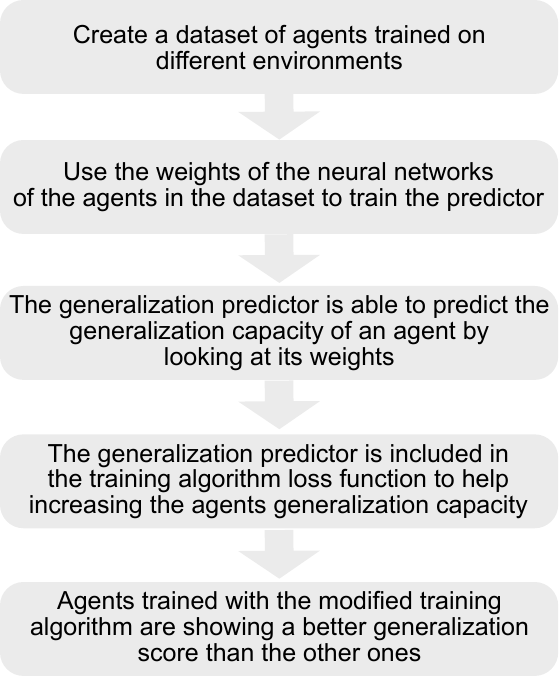}\hfill 
	\caption{In our approach, we first train a dataset of agents on different environments to serve as training inputs for the generalizability predictor. The generalizability predictor is then used to modify the loss function of the training algorithm to increase the generalization capacity of the agents trained with it.}
    \Description{}
	\label{overall_approach}
\end{figure}

The goal of our approach is to create a predictor able to estimate the generalizability of a given agent by only looking at its neural network's weights. We can then use this predictor to influence the loss function of the training algorithm in order to improve the generalizability of the agents during their training.
The different steps of the approach presented in this paper are described in Figure~\ref{overall_approach}
\subsection{Reinforcement Learning formulation}
Reinforcement learning is defined as a setup in which an agent learns from its interactions with its environment over discrete time steps, rather than learning from a predefined dataset.
The data used for the learning process are collected directly from the environment. Learning can be done directly during exploration or later using a replay-memory approach.
An environment used in Reinforcement Learning is formalized as a Markov Decision Process (MDP) as described by Puterman \cite{mlputerman1994}.
An MDP, noted $\mathcal{M}$, is composed of (i) a state space (all possible states of the agent), (ii) an action space (all possible actions to be taken by the agent), (iii) a transition function (setting the new state based on the action taken) and (iv) a reward function (indicating the reward received by the agent when taking a given action).
The state space is identified as $\mathcal S$. This state space can be continuous or discrete. 
The action space is defined as $\mathcal A=\{1, \ldots, N_{\mathcal A}\}$.
The transition function is noted $T:~\mathcal S \times \mathcal A  \to \mathbb{P}(\mathcal S)$. 
The reward function is noted $R:~\mathcal S \times \mathcal A \times \mathcal S \to \mathcal R$ where $\mathcal R$ includes all potential rewards in a range $[0,R_{\text{max}}]$.
For each trajectory, the agent starts in the initial state, sampled from a distribution of initial states, noted $b_0(s)$.
At each time step $t$, the agent selects an action $a_t \in \mathcal A$, taken from the list of available actions according to the current state of the environment. 
The policy $\pi:\mathcal S \rightarrow \mathbb{P}(\mathcal A)$: $a_t \sim \pi(s_t,\cdot)$ defines the action that is selected.
After taking the action, the agent moves to a new state $s_{t+1} \in \mathcal S$.
The agent will also receive a reward noted $r_t \in \mathcal R$. 
The objective of the agent is to find a policy $\pi^*$ that maximizes the expected cumulative discounted reward:
\[
\pi^* = \arg\max_{\pi} \; \mathbb{E}_{\pi} \left[ \sum_{t=0}^{\infty} \gamma^t r_t \right],
\]
where $\gamma \in [0,1)$ is the discount factor.

\subsection{Metric used to define generalizability}
\label{score}
The generalizability index based on the average reward gathered, noted as $\zeta$, is defined as the average of rewards $\mathcal{R}$ gathered in all the new environments $M_{i} \in \mathcal{M}$ in which the approach was tested:
\begin{center}
	$\zeta = \frac{\sum_{i=0}^{n} R_{M_{i}}}{n}$
\end{center}
\subsection{Creating a dataset of agents}
In order to be able to train a predictor capable of estimating the generalizability of an agent just by looking at its neural network's weights, it is needed to create a dataset of agents. This dataset is composed of each agent neural network's weights as data and their real generalization score as label.
The generalization score of each agent is defined by running each agent in a set of environments to which it has never been exposed and calculate the average reward gathered as defined in section \ref{score}.

\subsection{Predicting generalizability of a given agent}
Our approach is similar to the work from Unterthiner et al. \cite{tUnterthiner2021} who described that accuracy in Deep Neural Networks can be derived from the weights of the model. We adapt this approach to the specific context of Reinforcement Learning and to predict the generalization score on never seen environments.
Our approach is based on the intuition that generalizability of a neural network can be coded in patterns which can be found in its weights organization / distribution.
In order to analyze the weights of the agent neural network, we conduct experiments with two different model architectures. The first one is using a CNN which has proven to be efficient in recognizing patterns in pictures, for example, in the computer vision domain.(section \ref{analyzeCNN})
The second is inspired by the approach described by Unterthiner et al. \cite{tUnterthiner2021}, which is based on a DNN with an abstraction layer composed of statistical calculations extracted from the weights of the agent's neural network.(section \ref{analyzeDNN})
\subsection{Analyzing the weights with a CNN}
\label{analyzeCNN}
\begin{figure}[!htbp]
	\centering
	\includegraphics[width=7cm]{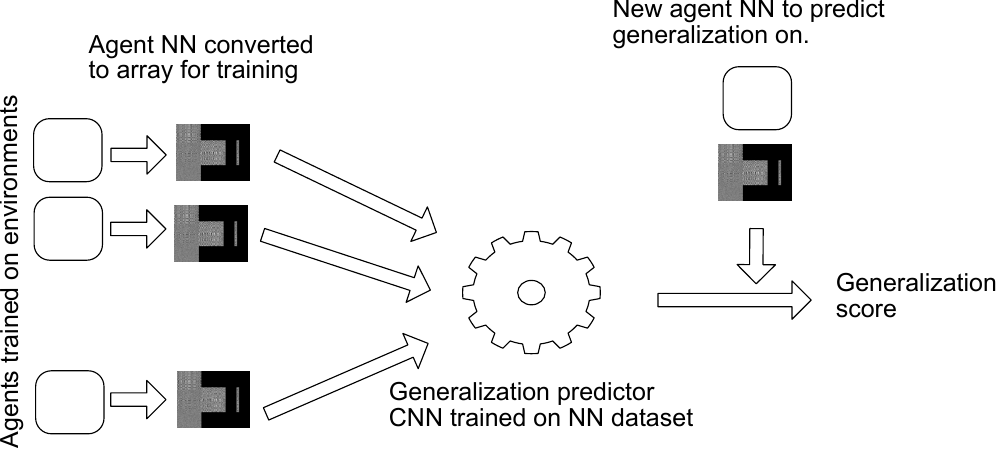}\hfill 
	\caption{From Agents to prediction of generalizability with a CNN, presenting the different steps from training the predictor and using it to predict the generalization score of a given agent based on its weights}
    \Description{}
	\label{overall_schema_CNN}
\end{figure}

We represent the agent's neural network as an array and train a convolutional neural network (CNN) to predict the agents' generalization scores.
CNNs are well-suited for identifying patterns in images or arrays, as they learn their filters during training. 
Our ideas for using a CNN is that the weights of a neural network can be presented as a data structure similar to the one of an image. We also assume that some patterns created between the weights in this data structure can be captured by the CNN filters, in the same way the CNN captures angles, shapes from an image.
As described in Figure \ref{overall_schema_CNN}, the CNN used as generalizability predictor is first trained with a dataset composed of the weights of the agents' neural networks.
For each agent we have calculated their generalization score, by running them on never seen environments. We then use this score as label during the training of the predictor. After the CNN is trained, it can be used to predict the generalization score of a new agent by submitting as input the neural network of this new agent. The output of the generalizability predictor being the predicted generalization score for this new agent.
\subsection{Analyzing the weights with a DNN}
\label{analyzeDNN}
\begin{figure}[!htbp]
	\centering
	\includegraphics[width=7cm]{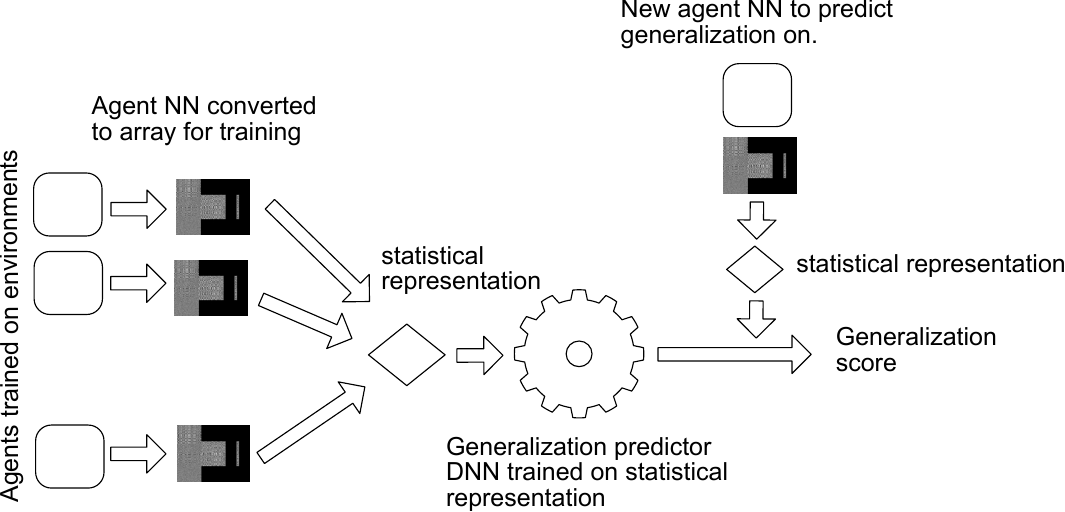}\hfill 
	\caption{From Agents to prediction of generalizability with DNN,presenting the different steps from training the predictor and using it to predict the generalization score of a given agent based on its weights}
    \Description{}
	\label{overall_schema_GBM}
\end{figure}

Instead of feeding directly the weights into the model, we compute a set of descriptive statistics for each layer's weights, transforming them into compact and informative input features for our prediction model.
A DNN can be defined as :
$y = F(x; \theta) = f^{(n)}(f^{(n-1)}(\dots (f^{(1)}(x)) \dots ))$
where y is the output of the model, $f^{(n)}$ represents each layer of the model, and x is the input of the model.
This setup introduces an additional abstraction layer, composed of weight-based statistical computations, compared to the CNN approach, which relies solely on raw weights or simple aggregate measures. This allows the DNN predictor to learn from a statistical representation of higher level of the agent's parameters.
The abstraction layer is being represented by $g(x)$, which means that the DNN with the abstraction layer is defined as: $y = F(g(x); \theta) = f^{(n)}(f^{(n-1)}(\dots (f^{(1)}(g(x))) \dots ))$
This concept of an abstraction layer to train a predictor is similar to the one proposed in a different setup by Unterthiner et al. \cite{tUnterthiner2021}.
The initial set of potential features assessed for our approach are the ones defined by Unterthiner et al. \citet{tUnterthiner2021}:
\begin{itemize}
\item Mean (Average)
\item Variance
\item Percentiles: 0th, 25th, 50th (Median), 75th, and 100th
\end{itemize}
Which means that if we consider $x$ being the raw weights of the agent neural network, the input $g(x)$ of the predictor will be:
\begin{center}
$g(x)=[M(x),V(x),P_{0}(x),P_{25}(x),P_{50}(x),P_{75}(x),P_{100}(x)]$
\end{center}
where:
\begin{itemize}
\item $M(x)$ is the mean of X 
\item $V(x)$ is the variance of X
\item $P_{n}(x)$ is the $n^{th}$ percentile of X
\end{itemize}
The difference in our approach is that we do a selection of the features to be used for training the predictor based on the pearson-correlation between the proposed features and the predicted generalization score and only select the most performing ones.
As illustrated in Figure \ref{overall_schema_GBM}, the DNN neural network used as generalizability predictor is trained with a dataset of abstraction layers calculated from the neural networks of agents. 
We also calculate the generalization score of each agent, by running them on never seen environments, to use it as label during training. When the training is completed, we can use the generalizability predictor to predict the generalization score of a new agent, by first calculating the abstraction layer from the new agent neural network weights and then submit it as input data to the generalizability predictor.

\subsection{Improving generalizability of the training algorithm}
To enhance the generalization capabilities of agents trained we introduce a modification to its original training algorithm loss function. 
\begin{figure}[!htbp]
	\centering
	\includegraphics[width=7cm]{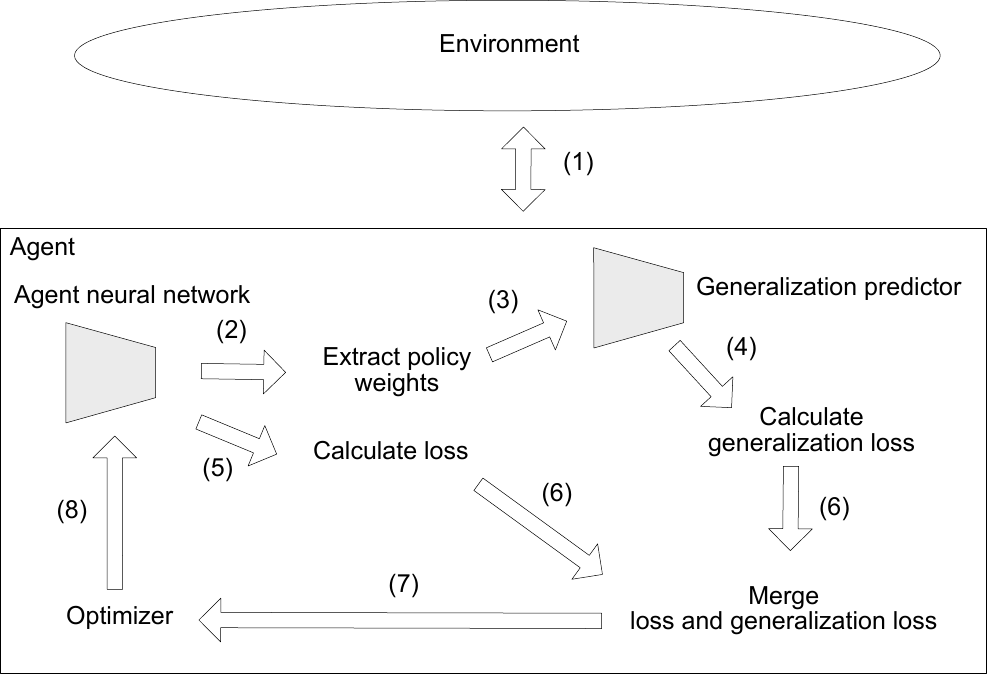}\hfill 
	\caption{Modification of the agent training algorithm using the trained predictor as loss function component to improve trained agent generalization capability (numbers represent order of actions}
    \Description{}
	\label{PPO_overview}
\end{figure}
Specifically, we add to the standard objective an additional term that incorporates a predicted generalization score based on the current policy neural network’s weights. As described in Figure \ref{PPO_overview} the actions occur in the following order : (1)~the agent interacts with the environment to gather data. (2)~the current policy (neural network weights) of the agent is extracted, the abstraction layer statistical elements are computed and (3)~fed to the generalizability predictor. (4)~the generalizability predictor predicts the generalization capacity of the agent neural network and the generalization loss. (5)~the loss is calculated and (6)~merged with the generalization loss to compute the overall loss which is used (7)~to feed the optimizer which will (8)~update the Agent neural network weights.
We modify the loss function by adding a term $G(\pi_{\theta})$, which represents the predicted generalization capacity of the policy $\pi_{\theta}$. This term is scaled by a coefficient $c_3$ to adjust its impact on the overall loss function. The inclusion of $G(\pi_{\theta})$ guides the optimization process toward policies that show improved generalizability to unseen environments. The optimization during the training cycle is done on both the performance on the training environment as well as the generalizability of the trained agent.

\subsection{Testing generalizability improvement}
The upgraded training algorithm is compared to the original one on the same environments, for the same number of training cycles and tested on the same number of never seen environment to determine its generalization score, offering a fair comparison.

\section{Experimental setup}
\label{experimental_setup}
\subsection{Training algorithm for agents and generalizability improvement}
Reinforcement Learning can be implemented using different approaches / algorithms, e.g. DDQN defined by Van Hasselt et al. \cite{Hvanhasselt2016}, Actor-Critic described by Konda et al. \cite{Vkonda1999}, and PPO presented by Schulman et al. \cite {jschulman2017}.
In the following experiments, all the agents part of the predictor training dataset, on both environments, as well as the agents used to show the improvement of our modified loss function approach are trained using the PPO algorithm.
The Proximal Policy Optimization (PPO) approach has been derived from and improves the actor-critic method by bringing more stability in the learning process.
The parameters, noted $w$ of a given policy $\pi_{w} (s,a)$ are updated to optimize $A^{\pi_{w}}(s,a)=Q^{\pi_{w}}(s,a)-V^{\pi_{w}}(s)$. 
The PPO algorithm adds a limit to policy changes that reduces variation at each training cycle, called clipping.
\subsection{Preparing the data}
\begin{figure}[!htbp]
	\centering
	\includegraphics[width=2cm]{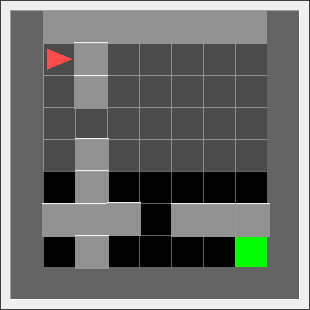}
	\includegraphics[width=2cm]{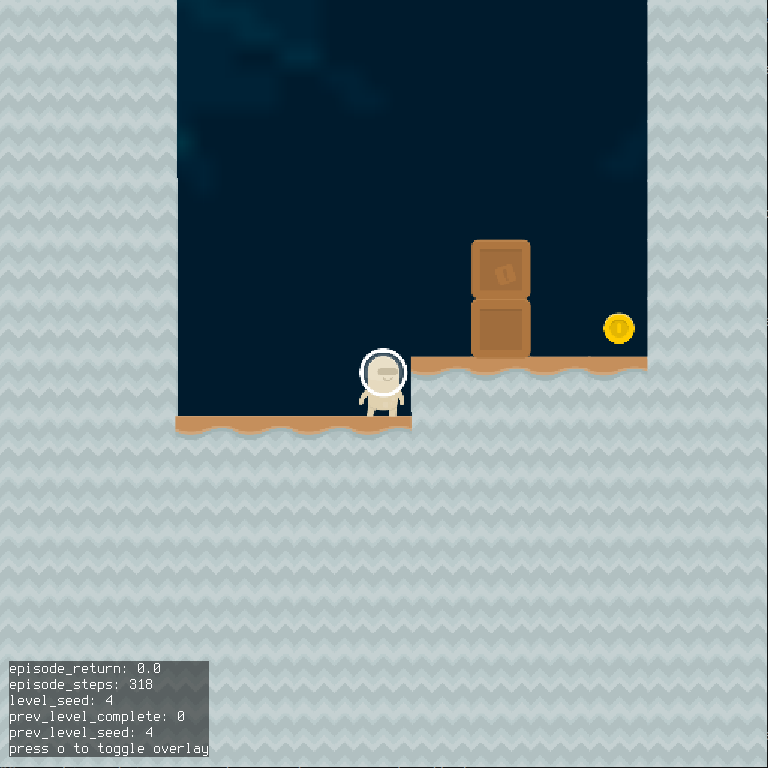}
	\caption{Minigrid SimpleCrossing and Coinrun environments used for training the predictors and testing the improved PPO algorithm}
    \Description{}
	\label{minigrid}
\end{figure}

In order to build the agent datasets to train the generalizability predictor used in the different experiments, we use the Minigird SimpleCrossing environment (Chevalier et al. \citep{gym_minigrid}) and the ProcGen coinrun environment (Cobbe et al. \citep{KCobbe2019}). (see Figure \ref{minigrid}).
Minigrid SimpleCrossing is a maze-type of environment where the agent needs to find a goal by navigating through 3 different areas separated by walls with one opening to cross them.
ProcGen coinrun is a platform-type game in which the agent should jump over stairs and objects to reach a goal (the coin) without falling in traps.
The environments are procedurally generated, meaning that their design is based on a seed, and ensure that each seed corresponds to a different version of the environment.
To ensure that the never seen environments are not included in the training dataset, we have used different seeds for them.
For the Minigrid setup, according to the reduced number of potential environments of this setup, random noise (adding a small random variation to the observation values returned by the environment) is added to them.
This noise modifies the observations returned and ensure that these observations were never part of the training dataset.
Each minigrid agent is trained on one environment for 2e6 steps (determined after testing for finding the most computationally efficient level of training while maintaining a proper accuracy) and tested for generalization score on 1000 never seen environments. We choose this approach in order to generate a wider sample of generalization scores. This dataset goes from agents which do not generalize at all to agents which generalize better. If we chose the option to train each agent on multiple environments, the generalization score will be more centered around the average generalization score not providing values like 0, and then restricting the scope of the generalizability predictor.
For the Coinrun setup, 3000 agents have been trained.
Coinrun offers a far wider number of potential environments and is created to ensure that two seeds will never give the same environment. It is then not needed to add the random noise on the observations returned by the environment, as it was the case for minigrid setup, to ensure that the never seen environment are really never seen.
Each coinrun agent is trained on a set of 20 environments for 1.5e5 steps (determined after testing for finding the most computationally efficient level of training while maintaining a proper accuracy) and tested for generalization score on 500 never seen environments. We chose to keep the standard approach of multi-levels training for the Coinrun environment, which was also offering a way to see if our approach works in a slightly different setup than for Minigrid.
For both approaches, each agent Neural Network is then converted to a 3 layers array (Figure \ref{agent_NN_image}) to be used to train the generalizability predictor.

\subsection{Agents model architecture}
\begin{figure}[!htbp]
	\centering
	\includegraphics[width=4cm]{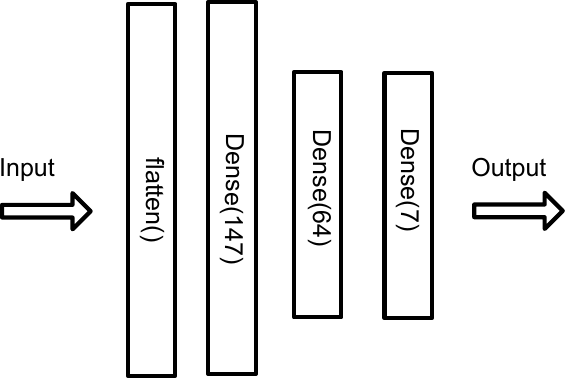}\hfill 
	\caption{Stable-baselines3 PPO MlpPolicy NN structure.In this setup, the PPO model is composed of 3 fully connected layers with 147, 64 and 7 neurons each.}
    \Description{}
	\label{agent_NN}
\end{figure}

For both types of environments, the models of the agents are trained using the vectorized version of the environment representation.
 Stable-baselines3 \cite{stable-baselines3} PPO algorithm with MlpPolicy is used as the neural network of each agent. This neural network is defined as a 3 layers network, as indicated in Figure~\ref{agent_NN}.
This means that the weights of the neural network of each agent is defined in 3 separate arrays, one per layer.

\begin{figure}[!htbp]
	\centering
	\includegraphics[width=6cm]{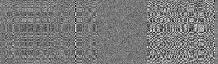}\hfill 
	\caption{Illustration of the Agents weights as a greyscale image. The weights have been normalized in values between 0 and 1 and then assigned a greyscale RGB value (R=G=B) with values between 0 and 255.}
    \Description{}
	\label{agent_NN_image}
\end{figure}

For the training of the CNN approach, the weights of the layers of the agent's neural network  are concatenated together in order to build one array in an image shape, while for the DNN the abstraction layer is composed of statistical calculations from each array separately.
The graphical view of the concatenated array converted to RGB image can be seen on Figure \ref{agent_NN_image}.
The image is shown here to give a visual representation of the Neural Network of an agent, but the training is conducted directly on the concatenated arrays and not on an image.

\subsection{Training the CNN and the DNN}
The CNN we use to predict generalizability is based on a set of convolutional and dense layers analyzing the concatenated array (Network details in the appendix). The network structure is inspired from vg16 introduced by K.Simonyan et al. \cite{KSimonyan2015}.
The output of the neural network is a regression layer, predicting the generalization score of a given agent based on its weights.
The CNN is trained with Adam optimizer and the mean squared error loss function.

For the DNN, the model is composed of a set of fully connected layers and the input values are based on an abstraction layer created by using statistical computations over the weights (Network details in the appendix).
The output of the neural network, similar to the CNN, is a regression layer, predicting the generalization score of a given agent based on its weights.
The optimizer as well as the loss function are identical to the ones used for the CNN.

\subsection{Modification of the PPO algorithm}
To use the predictor to impact the PPO loss function in order to trigger better generalizability during the training of the agent, the loss function is modified as follow, adding the $c_3 G(\pi_{\theta})$ term.
\\
\\
   $L_{\text{PPO}}(\theta, \phi) 
    =  - \mathbb{E}_t\left[\min\left(r_t(\theta)\hat{A}_t,\, 
        \text{clip}\left(r_t(\theta), 1-\epsilon, 1+\epsilon\right)\hat{A}_t\right)\right] \\
       + c_1 \mathbb{E}_t\left[
        \max\left((V_{\phi}(s_t) - V_t^{\text{target}})^2,\; 
        (V_{\phi}^{\text{clip}}(s_t) 
        - V_t^{\text{target}})^2\right)\right] \\
       - c_2 \mathbb{E}_t\left[H(\pi_{\theta}(\cdot|s_t))\right] \\
       + c_3 G(\pi_{\theta})$    
\\
\\Where:
\begin{itemize}
    \item $\hat{A}_t$: Advantage estimate at timestep $t$.
    \item $V_{\phi}(s_t)$: Current value function estimate.
    \item $V_{\phi_{\text{old}}}(s_t)$: Old value function estimate.
    \item $V_t^{\text{target}}$: Target value at timestep $t$.
    \item $H(\pi_{\theta}(\cdot|s_t))$: Entropy of the policy.
    \item $\epsilon$: PPO policy clipping parameter (typically $0.2$).
    \item $\epsilon_{\text{VF}}$: Value-function clipping parameter (often equal to $\epsilon$).
    \item $G(\pi_{\theta}$): Generalization score loss
    \item $c_1$: Coefficient for value function loss (typically $0.5$).
    \item $c_2$: Coefficient for entropy regularization (typically $0.01$).
    \item $c_3$: Coefficient for generalization score loss (typically $0.5$).
\end{itemize}
The PPO training loop in stable-baselines 3 \cite{stable-baselines3} is defined as indicated in the algorithm \ref{alg:upgradedPPO}.
In order to incorporate our new loss parameter in the overall PPO loss function, we make the following changes: the lines highlighted in green in the algorithm are added and the lines highlighted in orange are removed.
The generalization loss is defined as $L^{\text{gen}} \gets - G(\pi_{\theta})$ with $G(\pi_{\theta})$ being the estimated generalization score of the current policy given by the generalizability predictor. 

\begin{algorithm}
\caption{PPO stable-baselines3 training loop with highlighted changes}
\label{alg:upgradedPPO}
\begin{algorithmic}[1]
\STATE Initialize policy network $\pi_\theta$, value function $V_\phi$
\FOR{each iteration}
    \STATE Collect $n$ steps of experience using current policy $\pi_\theta$
    \STATE Compute returns and advantages using GAE:
    \STATE \hskip1em $\delta_t \gets r_t + \gamma V_\phi(s_{t+1}) - V_\phi(s_t)$
    \STATE \hskip1em $\hat{A}_t \gets \delta_t + \gamma \lambda \hat{A}_{t+1}$
    \STATE Store $(s_t, a_t, r_t, \hat{A}_t, \hat{R}_t)$
    \FOR{epoch = 1 to $n_{\text{epochs}}$}
        \STATE Shuffle data and divide into mini-batches
        \FOR{each mini-batch}
            \STATE Compute ratio: $r_t(\theta) \gets \frac{\pi_\theta(a_t|s_t)}{\pi_{\theta_{\text{old}}}(a_t|s_t)}$
            \STATE Compute policy loss:
            \STATE \hskip1em $L^{\text{clip}} \gets \min\left( r_t(\theta)\hat{A}_t,\; \text{clip}(r_t(\theta), 1 - \epsilon, 1 + \epsilon)\hat{A}_t \right)$
            \STATE Compute value loss: $L^{\text{vf}} \gets (V_\phi(s_t) - \hat{R}_t)^2$
            \STATE Compute entropy bonus: $L^{\text{ent}} \gets -\beta H[\pi_\theta(\cdot|s_t)]$
            \STATE \colorbox{green}{ Compute gen. loss: $L^{\text{gen}} \gets - G(\pi_{\theta})$}
            \STATE \colorbox{orange}{Total loss: $L \gets -L^{\text{clip}} + c_1 L^{\text{vf}} + c_2 L^{\text{ent}}$}
            \STATE \colorbox{green}{Total loss: $L \gets -L^{\text{clip}} + c_1 L^{\text{vf}} + c_2 L^{\text{ent}} + c_3 L^{\text{gen}}$}
            \STATE Update $\theta, \phi$ using gradients of $L$
        \ENDFOR
    \ENDFOR
    \STATE Update $\theta_{\text{old}} \gets \theta$
\ENDFOR
\end{algorithmic}
\end{algorithm}
When training with this updated PPO, the parameters of the generalizability predictor must not being updated during the backward propagation of the loss to the PPO neural network. The parameters of the generalizability predictor are then frozen, while the ones of the agent are modified by the optimizer. This prevents the optimizer to modify the generalizability predictor to reduce the Generalization loss by making it predict a more favorable generalization score. This would temper with the goal of the generalizability predictor. 
The parameter $c_3$ is set to 0.5 in our experiments. This coefficient is determined after testing multiple values between 0 and 2 and selecting the one offering the best generalization score increase to the agent.
\subsection{Testing the upgraded PPO algorithm}
In order to test the upgraded PPO algorithm, we train 30 agents with the standard PPO algorithm as well as with the Upgraded PPO algorithm for both Minigrid and Coinrun environments. Each agent is trained for $1e6$ steps and each $5e4$ steps, we test each version of the agent on 1000 never seen environments and calculate the average reward gathered.
The never seen environments reported in this part are different from the ones used for training / testing the predictor (different seeds), meaning that these environments remain really unseen. 
This approach gives us a fair comparison and enables us to spot the overall generalization score differences. This also allows us to test if one approach increases the generalization capacity of an agent earlier in the training process.
\section{Results}
\label{results}
We define the term of "proper signal" for the remaining part of this paper as a level of prediction where we can clearly  separate agents with low generalizability from agents with a good generalizability, as our goal is to use this signal as an input to improve the PPO algorithm. 
In term of Pearson correlation grading the following is generally accepted:
\begin{itemize}
    \item 0.8 to 1.0: very strong
    \item 0.5 to 0.79: strong
    \item 0.3 to 0.49: moderate
    \item 0.1 to 0.29: weak
    \item 0 to 0.1 : no correlation 
\end{itemize}
We consider that any Pearson correlation between the predicted generalization score and the labeled generalization score greater than 0.5 is a proper signal (as defined above).
\begin{figure}[!htbp]
	\centering
	\includegraphics[width=6cm]{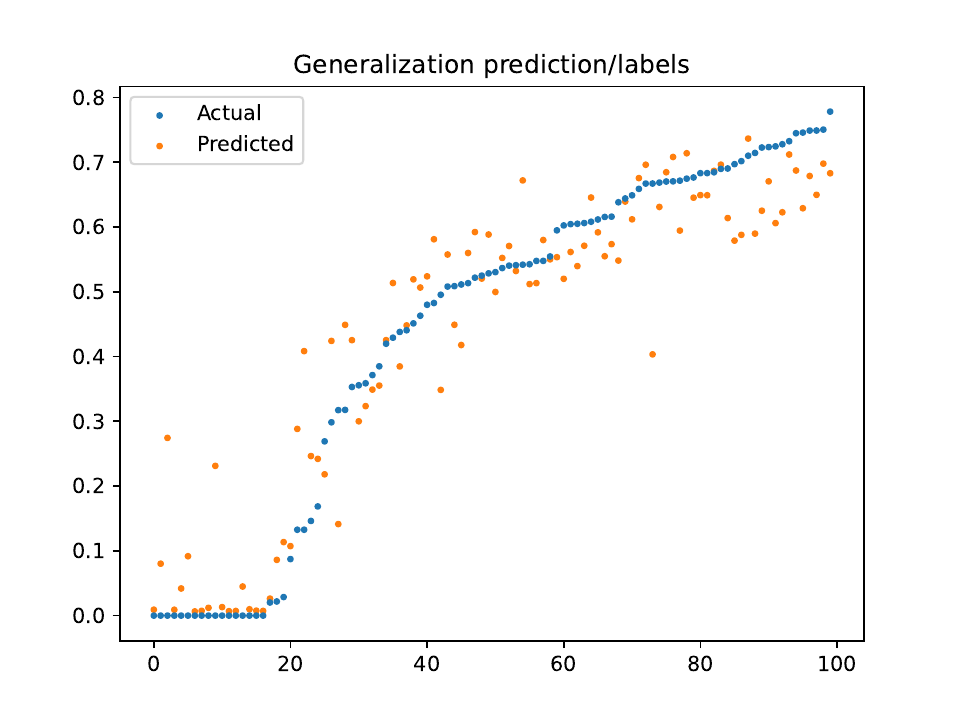}
	\caption{Minigrid - CNN prediction of generalizability based on weights. The y-axis is showing the generalization score (average of rewards) gathered  or predicted for each agent on never seen environment. Minigrid rewards are defined between 0 and 1. The x-axis indicates the number of the agent (0 to 100).}
    \Description{}
	\label{CNN_prediction_minigrid}
\end{figure}
\subsection{CNN model results}
\label{accuracy}
The CNN model is trained on the Minigrid dataset and Figure \ref{CNN_prediction_minigrid} shows how accurate the generalizability predictor is compared to the labeled data.
The CNN model can generate a proper signal, in the Minigrid context, for the generalizability of a PPO model, which will make it usable in an improved PPO training algorithm (Figure \ref{CNN_prediction_minigrid}). 

\begin{figure}[!htbp]
	\centering
	\includegraphics[width=6cm]{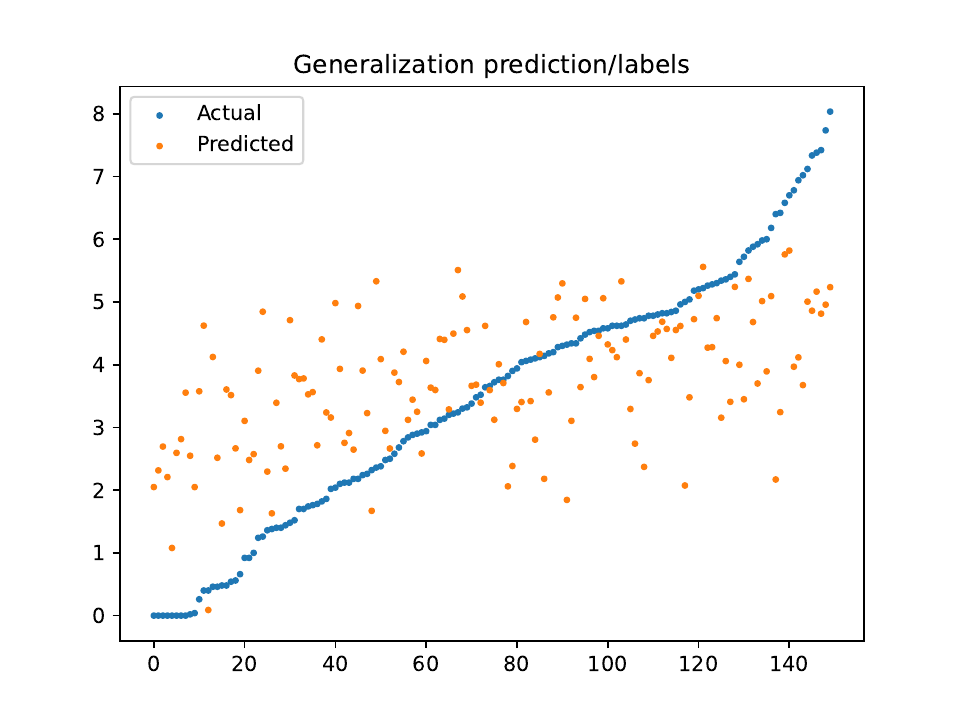}
	\caption{Coinrun - CNN prediction of generalizability based on weights. The y-axis is showing the generalization score (average of rewards) gathered or predicted for each agent on never seen environments. Coinrun rewards are defined between 0 and 10. The x-axis indicates the number of the agent (0 to 100).}
    \Description{}
	\label{CNN_prediction_coinrun}
\end{figure}

We can see a strong correlation between the predictions and the labeled data visually but also by calculating the Pearson correlation of the two datasets with a score of 0.941. (very strong)
The CNN model is also trained on the Coinrun dataset and the results are shown in figure \ref{CNN_prediction_coinrun}.
The CNN is not able to generate a proper signal (as defined above) for the generalizability of the agents in the Coinrun setup. Even if an indication of a potential signal seems to emerge, it is not strong enough to be used for improving the PPO loss function.

The weakness of the CNN prediction in this setup can be seen visually as well as with the calculation of the Pearson correlation of the predictions and the labeled data, having a score of 0.423. (moderate)
\begin{figure}[!htbp]
	\centering
	\includegraphics[width=6cm]{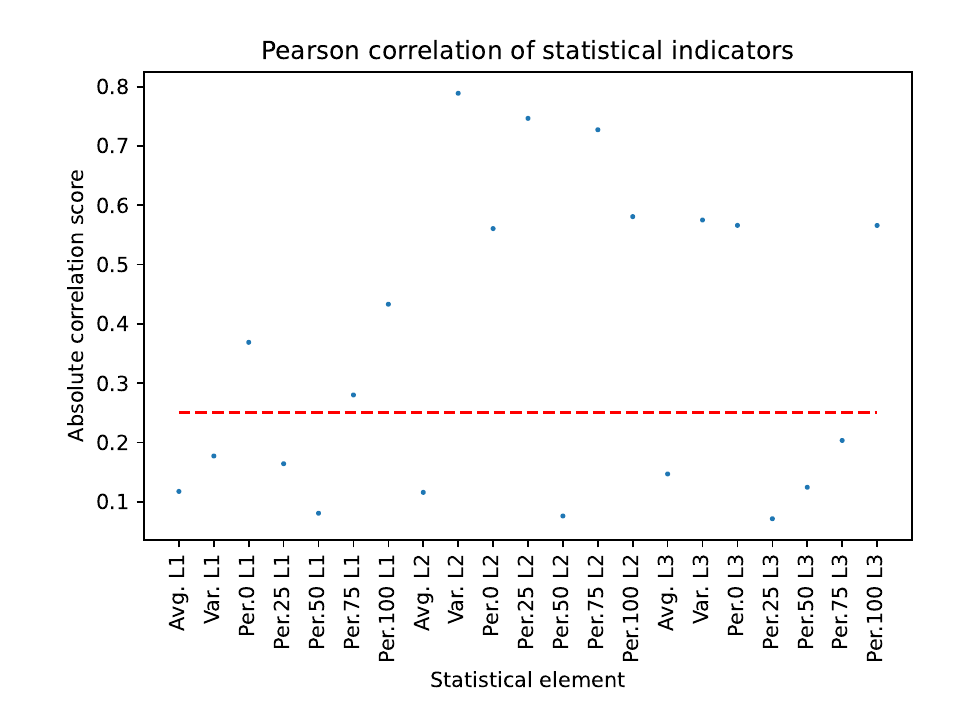}\hfill 
	\caption{Minigrid - Pearson correlation, indicating the extent to which each of the feature selected to train the predictor have an impact on the generalization score predicted. The red line indicates the threshold used in this setup to select the features. }
    \Description{}
	\label{pearson_correlation_minigrid}
\end{figure}
\subsection{DNN model results}
In order to build the abstraction layer of the DNN model for Minigrid and Coinrun setup,  we evaluate the Pearson correlation between each statistical measure and the agent's generalization score observed.
We want to ensure that we remove the metrics with the lowest correlation to the generalization score of the agent as they only brought noise and potentially reduced the signal quality. We select the threshold based on each setup by considering two factors, the correlation score being under 0.3 (weak) but also the remaining number of features and we adjust to the right level by running experiments. 

Only the statistics with the highest correlation, from which we are able to derive a signal, are retained for model input.
As shown in Figure \ref{pearson_correlation_minigrid}, the selected features with a significant correlation with the generalization score, in the Minigrid setup include:
$$\begin{array}{|l|c|r|} 
\hline \textbf{Layer 1} & \textbf{Layer 2} & \textbf{Layer 3} \\ 
\hline \text{Percentile 0} & \text{Variance} & \text{Variance} \\ 
       \text{Percentile 75} & \text{Percentile 0} & \text{Percentile 0} \\ 
       \text{Percentile 100} & \text{Percentile 25} & \text{Percentile 100} \\ 
       \text{} & \text{Percentile 75} & \text{} \\ 
       \text{} & \text{Percentile 100} & \text{} \\ 
\hline \end{array} $$\

\begin{figure}[!htbp]
	\centering
	\includegraphics[width=6cm]{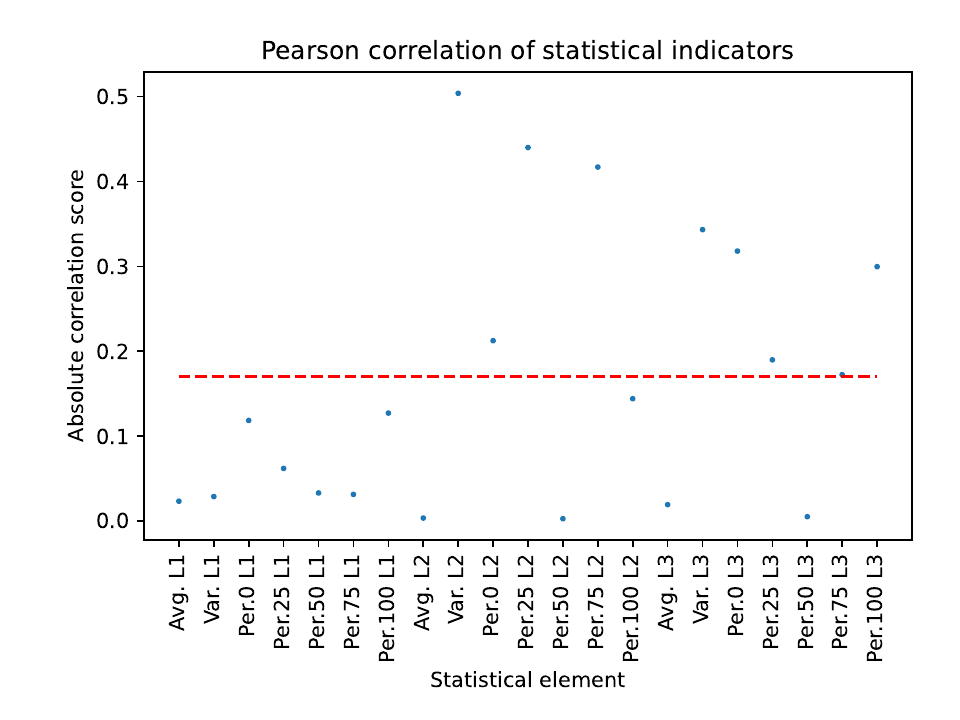}\hfill 
	\caption{Coinrun - Pearson correlation, indicating how much each of the feature selected to train the predictor have an impact on the generalization score predicted. The red line indicates the threshold used in this setup to select the features. }
    \Description{}
	\label{pearson_correlation_coinrun}
\end{figure}

\begin{figure}[!htbp]
	\centering
	\includegraphics[width=6cm]{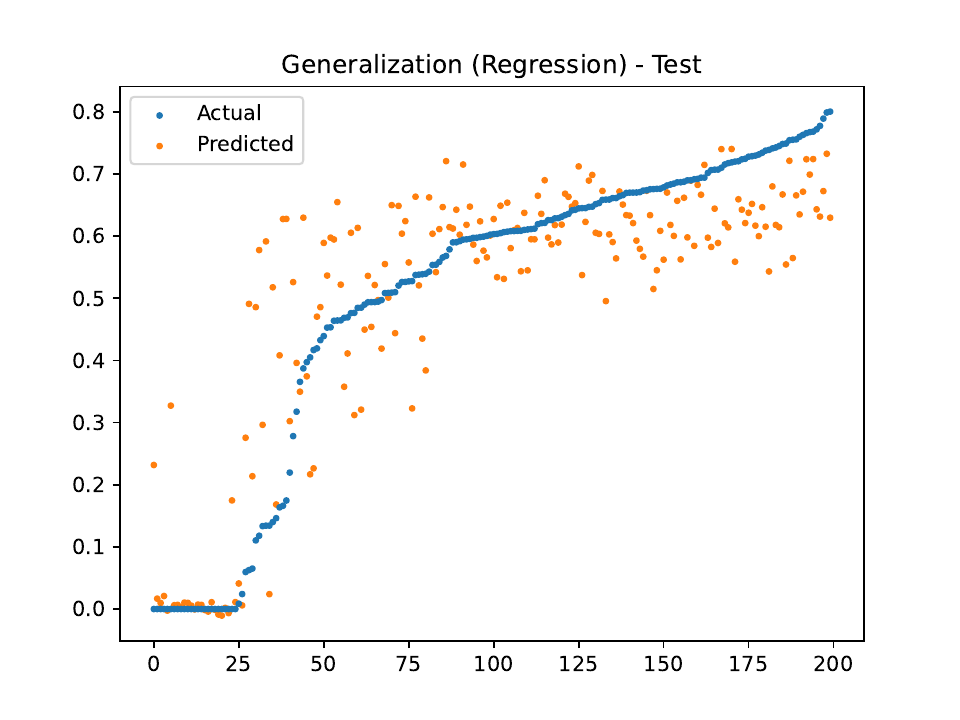}
	\caption{Minigrid - DNN prediction of generalizability. The y-axis is showing the generalization score (average of rewards) gathered or predicted for each agent on never seen environments. Minigrid rewards are defined between 0 and 1. The x-axis indicates the number of the agent.(0 to 200)}
    \Description{}
	\label{DNN_prediction_minigrid}
\end{figure}

\begin{figure}[!htbp]
	\centering
	\includegraphics[width=6cm]{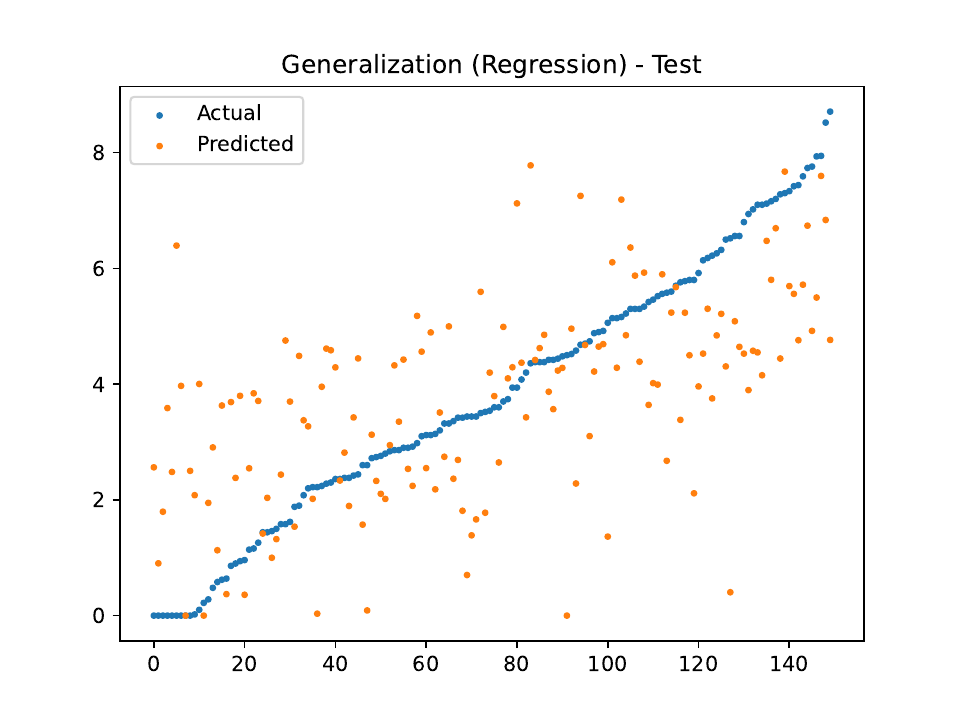}
	\caption{Coinrun - DNN prediction of generalizability. The y-axis is showing the generalization score (average of rewards) gathered or predicted for each agent on never seen environments. Coinrun rewards are defined between 0 and 10. The x-axis indicates the number of the agent.(0 to 150)}
    \Description{}
	\label{DNN_prediction_coinrun}
\end{figure}

In the Coinrun setup, the selected features (Figure \ref{pearson_correlation_coinrun}) are quite different from the minigrid environment, which may be linked to the complexity of the environment. The ones included are:
$$\begin{array}{|l|r|} 
\hline  \textbf{Layer 2} & \textbf{Layer 3} \\ 
\hline  \text{Variance} & \text{Variance} \\
        \text{Percentile 0} & \text{Percentile 0} \\ 
        \text{Percentile 25} & \text{Percentile 25} \\ 
        \text{Percentile 75} & \text{Percentile 75} \\ 
        \text{} & \text{Percentile 100} \\ 
\hline \end{array} $$\
The DNN model is trained on the Minigrid dataset and figure \ref{DNN_prediction_minigrid} shows how well it can predict generalizability compared to labeled data.
In the Minigrid context, the DNN is able to generate a proper signal (as defined in section \ref{accuracy}) which is confirmed visually, as well as by calculating the Pearson correlation of the predictions and the labeled data used for testing, achieving a score of: 0.885. (very strong)
The DNN model is trained on the Coinrun dataset and its performance for predicting generalizability can be seen on figure \ref{DNN_prediction_coinrun}.

The DNN is also able to generate a proper signal (as defined in section \ref{accuracy}) in the Coinrun context. It is confirmed visually, but also by calculating the Pearson correlation score between the predictions and the labeled data used for testing. The score , 0.568, (strong) is lower than for Minigrid setup, but still shows a good correlation which makes it a strong enough signal to be used in our improved PPO approach.
This means that the DNN approach is capable of finding a clear signal in both approaches, leading to strong and very strong correlation.

\subsection{Upgraded training conditions}
The DNN approach which proves to be quite consistent in terms of results between the two environments, is also faster to train, and lighter in term of computing resources as detected when running the experiments (both in training and inference). It is an approach of choice for predicting generalizability in the context of a model training loop, where we need inference to be fast. It is then the one we choose for our proposed modification of the PPO algorithm.
\begin{figure}[!htbp]
	\centering
	\includegraphics[width=6cm]{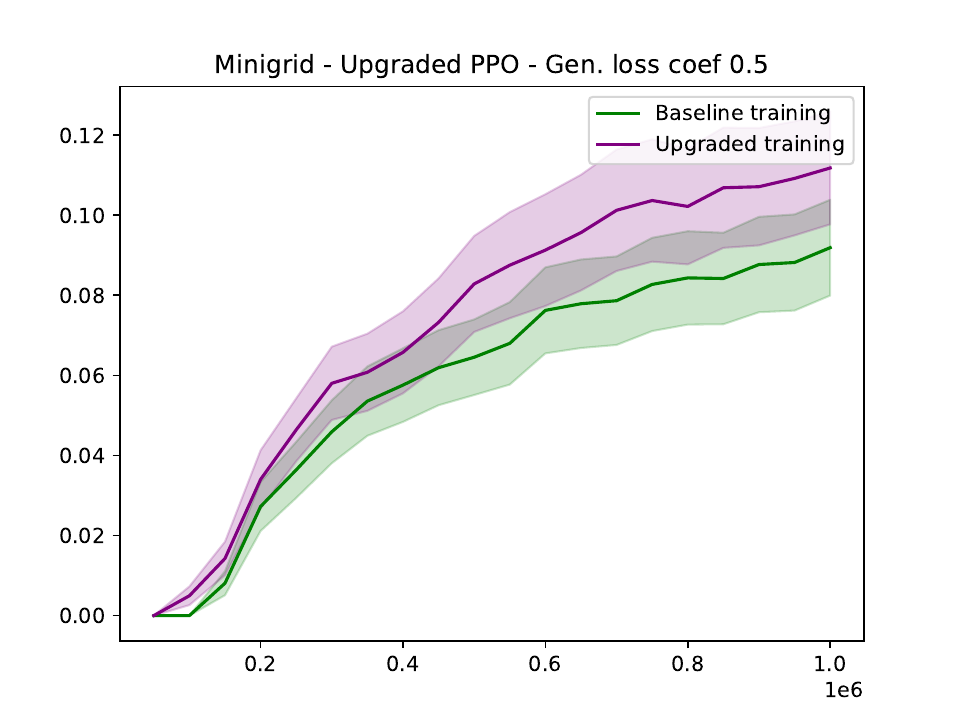}
	\caption{Minigrid - Upgraded PPO algorithm performance over the 30 test environments with standard error. The y-axis indicates the generalization score (average rewards, max = 1) gathered on never seen environments from the standard (green) and upgraded PPO (purple) algorithms and the x-axis indicates the number of training steps. This is based on the average of 30 agents being trained (line) and the associated standard error (shadow around the line)}
    \Description{}
	\label{upgradedPPO_minigrid}
\end{figure}

\begin{figure}[!htbp]
	\centering
    	\includegraphics[width=6cm]{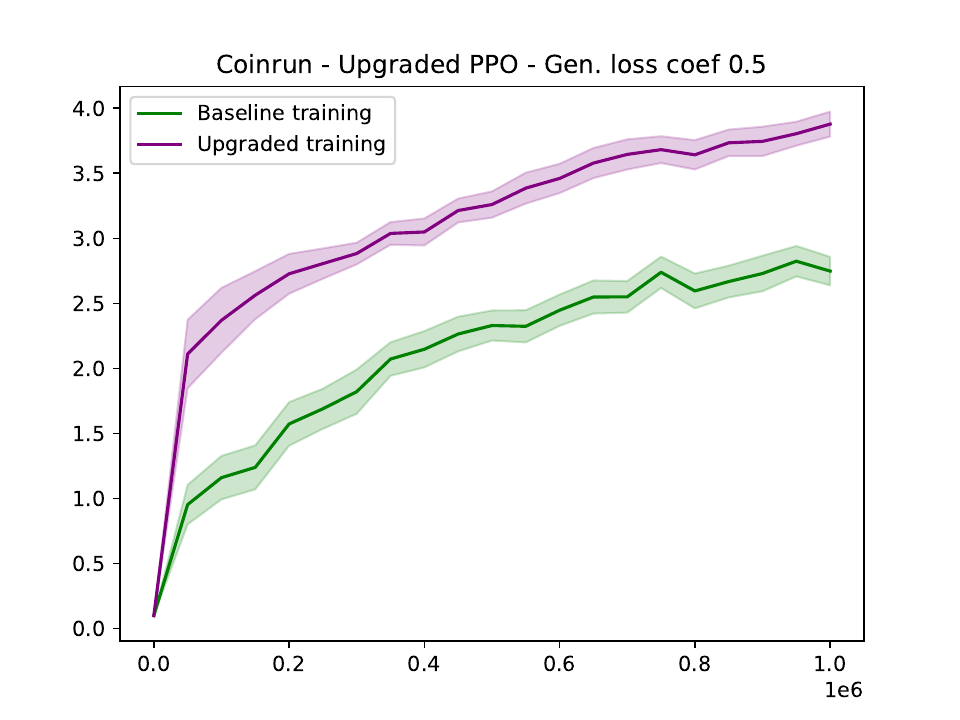}
	\caption{Coinrun - Upgraded PPO algorithm performance over the 30 test environments with standard error. The y-axis indicates the generalization score (average rewards, max = 10) gathered on never seen environments from the standard (green) and upgraded PPO (purple) algorithms and the x-axis indicates the number of training steps. This is based on the average of 30 agents being trained (line) and the associated standard error (shadow around the line)}
    \Description{}
	\label{upgradedPPO_coinrun}
\end{figure}

Following the upgrade of the PPO algorithm, using generalizability prediction impact on the loss function to improve its training, we can see that the upgraded PPO algorithm is superior to the standard PPO algorithm at all times.
On average, the upgraded PPO generalizes better and earlier than the standard PPO algorithm.
The chart representing both training versions performance shows the average generalization score as well as the associated standard error. (Figure \ref{upgradedPPO_minigrid} and Figure \ref{upgradedPPO_coinrun})
We can see that the modification of the loss function has a direct impact on the generalizability of the agents being trained and that we can achieve multiple objectives, performance and generalizability in our case, by adding additional terms to the loss function.

\section{Discussion / conclusion}
\label{discussion_conclusion}
We show that the weights configuration of a reinforcement learning agent using a neural network is clearly related to its ability to generalize to unseen environments. Building on the methodology introduced by Unterthiner et al. \cite{tUnterthiner2021}, we develop a predictive model capable of estimating an agent’s generalization score based on its neural network's weights, and test it in the context of PPO and reinforcement learning.
Leveraging this predictive model, we propose an improved version of agent training algorithm by incorporating a generalizability term into the loss function in order to force the neural network not only to achieve great performance in the environments in which it is trained, but also to increase its generalizability to never seen environments. 
Our results show that agents trained with this new training objective have a better generalizability.
We also show that having a better understanding of the weight structure of the neural networks or defining models capable of extracting these patterns can help define and increase some specific traits in the agent. This leads us to consider not only the neural networks and their weights as a black box, but more as an area to explore further in order to understand how its organization impacts different characteristics of a given agent.

Improving the generalizability of agents can lead to interesting behaviors for real-world applications, especially in domains like healthcare, where agents must often operate in slightly different environments (for example, different patient profiles). 
In this specific experimental setup, we focus on the PPO algorithm, but our approach can be extended to other types of RL/non-RL training algorithms as the additional generalizability is created by the impact of the predictor on the training loss function, which is a common aspect of most of the Reinforcement Learning training algorithms.

Overall, our approach also reduces significantly the computation power needed to assess an agent generalizability compared to the approach consisting of running each agent in the never seen environments.

\bibliographystyle{ACM-Reference-Format} 
\bibliography{aamas}

\end{document}